\documentclass[runningheads]{llncs}
\usepackage{graphicx}
\usepackage{amsbsy}
\begin{document}
	
	\title{Machine learning applications in time series hierarchical forecasting}
	
	\author{Mahdi Abolghasemi\inst{1}\orcidID{0000-0003-3924-7695} 
		Rob J Hyndman\inst{2}\orcidID{0000-0002-2140-5352}\\
		Garth Tarr\inst{3}\orcidID{0000-0002-6605-7478}
		Christoph Bergmeir\inst{4}\orcidID{0000-0002-3665-9021} }
	\authorrunning{M. Abolghasemi et al.}
	\institute{Monash University, Melbourne, VIC 3800, Australia
		\email{mahdi.abolghasemi@monash.edu} \and
		Monash University, Melbourne, VIC 3800, Australia
		\email{rob.hyndman@monash.edu} \and
		University of Sydney, Sydney, NSW 2006, Australia
		\email{garth.tarr@sydney.edu.au}
		%\url{http://www.springer.com/gp/computer-science/lncs} 
		\and
		Monash University, Melbourne, VIC 3800, Australia
		\email{christoph.bergmeir@monash.edu}}
	\maketitle          
	\begin{abstract}
		Hierarchical forecasting (HF) is needed in many situations in the supply chain (SC) because managers often need different levels of forecasts at different levels of SC to make a decision. Top-Down (TD), Bottom-Up (BU) and Optimal Combination (COM) are common HF models. These approaches are static and often ignore the dynamics of the series while disaggregating them. Consequently, they may fail to perform well if the investigated group of time series are subject to large changes such as during the periods of promotional sales. We address the HF problem of predicting real-world sales time series that are highly impacted by promotion. We use three machine learning (ML) models to capture sales variations over time. Artificial neural networks (ANN), extreme gradient boosting (XGboost), and support vector regression (SVR) algorithms are used to estimate the proportions of lower-level time series from the upper level. We perform an in-depth analysis of 61 groups of time series with different volatilities and show that ML models are competitive and outperform some well-established models in the literature.
		\keywords{Hierarchical forecasting \and group time series \and demand variations \and dynamic disaggregation \and machine learning.}
	\end{abstract}
	%
	%
	%

	% % % % % % % % % % % % % % % % % % % % % % % % % % % % % % % % % % % % % % % % %

	\section{Introduction}
	Hierarchical time series represent multiple time series that are hierarchically organized at different levels and can be aggregated and disaggregated at different levels with respect to various features such as location, size, and product type \cite{hyndman2011optimal}. 
	
	Forecasts can be generated either individually or via different HF approaches. If we forecast series independently, they may not add up properly because of the hierarchical constraints, so we would need to reconcile forecasts. The HF can potentially improve the forecast accuracy in comparison with the direct forecast if the HF model is chosen correctly \cite{nenova2016determining}.
	
	TD, BU, and COM approaches are the most common HF models in the literature. The efficacy of these models depends on the time series feature, the level of forecasting, forecasting horizon, and the structure of the hierarchy. We may consider these variables when choosing the optimal HF model \cite{nenova2016determining}.

	Different internal and external variables can change the dynamic of the series and affect the HF models performance. For example, promotion is a popular event in the fast-moving consumer goods (FMCG) industry to increase sales and drive major changes in the underlying demand behaviour \cite{Nikolopoulos2015}. Promotion can enhance sales of different entities at different levels of the hierarchy with different scales. That is, different entities of the hierarchy may experience different levels of uplifts in sales. As a result, sales proportions of each node from the total sales in the hierarchy varies over time and across different levels of the hierarchy. Therefore, we require a dynamic model for HF that can incorporate the impact of promotion and capture sales variations.  

	We propose to use ML models to disaggregate time series. We utilise three ML models to forecast the disaggregation factor, whereby the ANN, XGBoost and SVR models are used. Although ML models have been used frequently in demand forecasting \cite{makridakis2018,min2010artificial}, there is no empirical study to apply ML models in HF. We use a middle-out (MO) approach where forecasts are generated at the middle level. Then, we aggregate the middle-level forecast to the top level and disaggregate to the lower levels. We use a numerical approach on a group of sales time series of a company in Australia with a two-level SC. We obtain forecasts at different levels of the hierarchy which provide meaningful insights for practitioners to make a decision at different levels of SC  such as production planning, demand planning, and inventory control.  We also tackle the problem of demand volatility due to promotion in hierarchical context. Promotion impact on sales is an interesting open problem for researchers. 

	The remainder of the paper is organised as follows. Section \ref{literaturehierarchy} reviews the literature. Section  \ref{Methodologyhier} explains the hierarchical structure of the investigated data-set. Section 4 reviews the current HF models in the literature. Section \ref{dynamicproposalhier} discusses our ML approach. Empirical results and conclusion are drawn in Sections \ref{hierarchicalresults} and  \ref{hierarchicalconclusion}, respectively.

	\section{Literature review} \label{literaturehierarchy}
	
	Hierarchical time series can be built in different ways according to the different feature in the hierarchy. 
	The HF models benefit from some time series feature and techniques to improve  forecasting accuracy and generate efficiencies. There are many debates about the accuracy of HF models but there is no general consensus about their accuracy \cite{Fliedner1999,gross1990disaggregation}. Different factors such as the forecasting techniques, demand correlation between different series, the level of the hierarchy, and forecasting horizon may contribute to favouring one model to other models \cite{Fliedner1999,Fliedner2001,gross1990disaggregation}.

	The HF techniques are generally either from top of the hierarchy to the lower levels or from the bottom levels to higher levels. BU approach can be accurate at the bottom level. The downside of the BU model is that it can be difficult to model and forecast when the bottom level series are subject to noise. In the case of a large hierarchy, BU can be labour intensive and computationally expensive. More importantly, information may get lost when aggregating the lower-level series to forecast the upper-level series \cite{dangerfield1992top}.
	While there is no conclusive agreement on the performance of the HF models, often TD models are more appropriate for higher-level forecasts when strategic plans and decisions such as budgeting are required to be made. However, the BU model suits for operational decisions at the lower levels such as logistics and production planning \cite{kahn1998revisiting}. TD models generally require fewer forecasts. TD models are often accurate at the top level; however, their accuracy deteriorates by moving towards the lower levels of the hierarchy. This is because estimating the accurate proration factor is a challenging problem and allocating a disaggregation technique potentially introduce some errors to the forecasts.

	The problem of TD and BU models is that they start from the opposite ends of the hierarchy and their forecasts at the start of their point are more accurate. They do not use forecast from other levels of the hierarchy. Therefore, they lose some useful information \cite{pennings2017integrated}. 
	MO is a common approach in practice that tries to take the advantages of both the BU and the TD \cite{hyndman2011optimal}. This approach is essentially a combination of BU and TD where an intermediate level is chosen and forecasts are disaggregated to lower-level series and aggregated to higher-level series.
	
	COM is the other HF model that forecasts at all levels and combines them with a regression model to find the final reconciled forecasts \cite{hyndman2011optimal}. COM approach has a number of advantages over the BU and TD approaches. First, it considers the forecast at all levels of the hierarchy in contrast to the BU that only uses the lower-level series and the TD method that only uses the top level of the hierarchy. Second, we can use any kind of method to forecast demand and combine them with the COM approach. That is, any statistical model, ML model, or even judgemental forecasts can be used to generate base forecasts and reconciled with the optimal combination. However, this approach has its own pitfalls. This approach makes some assumptions in estimating the forecasting errors of the variance-covariance matrix that may be unrealistic \cite{pennings2017integrated}. Moreover, it neglects the sales dynamic and generates forecasts without considering the time impact \cite{Krick2012}.

	The other important factor in choosing the appropriate HF model is the level at which we are forecasting. Higher-level series can have higher accuracy as they are aggregated \cite{syntetos2016supply,zotteri2007model}. Demand characteristics also play an important role in choosing the most suitable HF model \cite{gross1990disaggregation}. The performance of HD models such as TD and BU have been investigated when demand follows AR(1), MA(1), and intermittent pattern \cite{moon2013development,widiarta2007effectiveness,widiarta2008forecasting}. However, promotion impact has not been investigated in HF models. Since promotion impact on sales my last over a number of periods it is crucial to consider a dynamic model rather than a static model when aggregating/disaggregating series \cite{Krick2012}. We use ML approach to address this problem. It is notable that there is no empirical study to investigate the ML models performance in HF.

	% Bayesian approach has also been used for dynamic proration in the HF \cite{park2014variational}.
	
	\section{Data}\label{Methodologyhier}
\begin{figure}
		\caption{Hierarchical structure of time series}
		\includegraphics[scale=0.26]{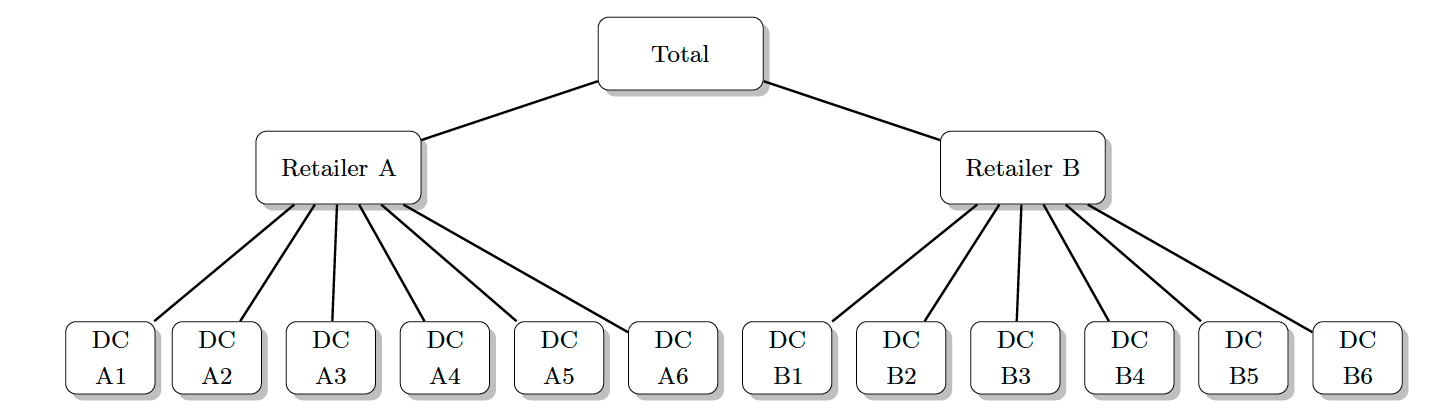}
		\label{htsstructure}
	\end{figure}

	We have gathered weekly sales for 61 groups of sales time series and prices across different locations. In total, we have more than 90,000 observations. Sales time series are for cereal and breakfast products for two major retailers across different states in Australia.  Fig. \ref{htsstructure} shows the hierarchical form of our case study and time series. The top-level represents the manufacturer total sales. There are two retailers at the middle level (level one). These retailers have different demand patterns. There are 12 different DCs at the bottom level (level two). These DCs have a similar demand pattern to their retailers (parent node) in terms of the promotion time. Retailers sale are correlated, and we can show that with a different structure of the investigated group of time series. However, we keep the hierarchical structure depicted in Fig. \ref{htsstructure} as it is useful for both our investigated case and research purposes.

	\section{Hierarchical forecasting models}\label{models}
	In this section, we discuss TD, BU, and COM approaches as three well-established HF models. The following indexes, notations, and parameters are used throughout this paper:\\
	%$x$: Series of the hierarchy\\
	$m_i$: Total number of the series for level $i$\\
	$k$: Total number of the levels in hierarchy\\
	$n$: Number of the observations\\
	%$x_t$: The $t$th observation of series x\\
	$\textbf{Y}_{i,t}$: The vector of all observations at level $i$\\
	$\mathbf{\hat{Y}}_{i,t} (h)$: $h$-step-ahead forecast at level $i$\\
	$\mathbf{Y}_t$: A column vector including all observations\\
	$\mathbf{\hat{Y}}_n (h)$: $h$-step-ahead independent base forecast of total based on $n$ observations \\
	$Y_{x,t}$: The $t^{th}$ observation of series $Y_x$\\
	$\hat{Y}_{x,n} (h)$: $h$-step-ahead independent base forecast of series $Y_x$ based on $n$ observations\\ 
	$\mathbf{\tilde{Y}}_n (h)$: The final revised forecasts\\
	
	These models can be shown with a unified structure $\mathbf{Y}_t$ = $\mathbf{S}$$ \mathbf{Y}_{k,t}$, where $\mathbf{S}$ is a summing matrix of order $m$ $\times$ $m_k$ that aggregates the bottom level series. 
	
	The general form of hierarchical models is $\mathbf{\tilde{Y}}_n (h)= \mathbf{S} \mathbf{P} \mathbf{\hat{Y}}_n(h)$, where $\mathbf{P}$ is a matrix of order  $m$ $\times$ $m_k$ and its elements depends on the type of the HF method that we use.  We use `\textit{hts}'  package in R for modelling of this section \cite{htspackage}.

	\subsection{Bottom-Up}\label{buhier}
	The BU approach forecasts at the bottom level and sums them up to find the forecast for higher levels. 
	In this approach, $\mathbf{P}= [\mathbf{0}_{m_k \times (m - m_k)}| \mathbf{I}_{m_k}]$, where $\mathbf{0}_{i \times j}$ is a $i \times j $ null matrix. In this case, $\mathbf{P}$ obtains the bottom level forecasts and combines them with the summing matrix $\mathbf{S}$ to generate the final forecasts of the hierarchy.
	
	\subsection{Top-Down}\label{tdhier}
	
	In the TD approach, forecasts are generated at the top level and disaggregated to the lower levels with an appropriate factor. 
	
	Gross and Sohl investigated 21 different disaggregation methods for TD models \cite{gross1990disaggregation}. They concluded that Equations \ref{td1} and \ref{td2} indicate two disaggregation methods that give reasonable forecast at the bottom level. 
	
	\begin{equation} \label{td1}
	p_j = \frac{\sum_{t=1}^{n}\frac{Y_{j,t}}{Y_t}}{n} ~~~~~~~~  j=1, ..., m_k 
	\end{equation}

	\begin{equation}\label{td2}
	p_j = \frac{\frac{\sum_{t=1}^{n} y_{j,t}}{n}}{\frac{\sum_{t=1}^{n} y_t}{n}} ~~~~~~~~  j=1, ..., m_k
	\end{equation}
	
	In the Equation \ref{td1} each proportion $p_j$ reflects the average of the historical proportions of the bottom level series ${Y_{j,t}}$.
	In the Equation \ref{td2} each proportion $p_j$ reflects the average of the historical value of the bottom level series ${Y_{j,t}}$ relative to the average value of the total aggregate ${Y_t}$. 
	
	These will form the $\mathbf{p}$ vector, $\mathbf{p} $=[$p_1$, $p_2$, $p_3$, ..., $p_{m_k}$], and $\mathbf{P}$ matrix as $\mathbf{P}= [~\mathbf{p} ~| ~\mathbf{0}_{m_k \times (m - 1)}]$. For the TD models, $\mathbf{P}$ disaggregate the forecast at the top level to the lower levels.

	\subsection{Top-Down forecasted proportions}\label{tdfphier}
	The above-mentioned TD approaches are static and can be biased over the horizon.
	Athanasopoulos et al. proposed top down forecasted proportions (TDFP) model for disaggregating the top level forecasts based on the foretasted proportions of lower level series rather than the historical proportions \cite{athanasopoulos2009hierarchical}. In this method, $p_j = \prod_{i=0}^{k-1} \frac{\hat{Y}^{(i)}_{j,n} (h)} {\sum (\hat{Y}^{(i+1)}_{j,n} (h))}$, for $j=1, ..., m_k$, where $\hat{Y}^{(i)}_{j,n} (h)$ is the $h$-step ahead forecast of the series that corresponds to the node which is $i$ levels above $j$, and $\sum\hat{Y}_{i,n} (h)$ is the sum of the $h$-step ahead forecasts below node $i$ that corresponds directly to the node $i$. These will form the $\mathbf{p}$ vector, $\mathbf{p} $=[$p_1$, $p_2$, $p_3$, ..., $p_{m_k}$] and $\mathbf{P}$ matrix as $\mathbf{P}= [~\mathbf{p} ~| ~\mathbf{0}_{m_k \times (m - 1)}]$. The downside of the TDFP model is that it generates biased forecast even if the base forecasts are unbiased. This is a disadvantage of all TD models \cite{athanasopoulos2009hierarchical}.

	\subsection{Optimal combination}\label{optimalhier}
	
	COM model uses a completely different approach for the HF. This approach uses a regression model to combine the forecasts at all levels, $\mathbf{\hat{Y}}_n (h)= \mathbf{S}  \pmb{ \beta_n} (h)+  \mathbf{\epsilon_t}, ~~ \pmb{\epsilon_t} \sim N(\mathbf{0}, \mathbf{W}_h )$, where $\pmb{\beta_n} (h) = E[\mathbf{\hat{Y}}_{k,n}(h) | \mathbf{Y_1}, ... , \mathbf{Y_n}]$ is the unknown mean of the base forecasts of the bottom level $k$. 
	
	Hyndman et al. showed that if $\mathbf{W}_h$ is known, then we can use generalised least square to compute the minimum variance unbiased estimate value of $\mathbf{\hat{Y}}_{k,t}(h)$ as $\pmb{\hat \beta_n}(h)=\mathbf{(S'W^\dagger _hS)}^{-1}\mathbf{S' W}^{\dagger}_h \mathbf{\hat{Y}_n}(h)$, where $\mathbf{W^\dagger _h}$ is the generalised inverse of $\mathbf{W_h}$ \cite{hyndman2016fast}. However, $ \mathbf{W}_h$ is not known and sometimes is impossible to estimate for large hierarchies. Different approximation methods can be used to estimate it. They showed that weighted least square can be used to estimate $\pmb{\hat \beta_n}(h)=\mathbf{(S'\Lambda _hS)}^{-1}\mathbf{S' \Lambda}_h \mathbf{\hat{Y}_n}(h)$, where $\mathbf{\Lambda_h}$ is a diagonal matrix with elements equal to the inverse of the variances of $\mathbf{\epsilon_h}$. Hence, $\textbf{P}=\mathbf{(S'W^\dagger _hS)}^{-1}\mathbf{S' W}^{\dagger}_h$ and the variance of revised forecasts is $Var[\mathbf{\tilde{Y}_n (h)}]= \mathbf{S}(\mathbf{S'W_h^{\dagger }S})^{-1}\mathbf{S'}$. We call this COM-WLS in this paper.
	
	Wickramasuriya et al. showed that for any $\mathbf{P}$ such that $\mathbf{SPS=S}$, the covariance matrix of the $h$-step ahead reconciled forecast errors is given by $\mathbf{V}_h=Var[\mathbf{y_{t+h}} - \mathbf{\tilde{Y}_t (h)}]= \mathbf{SPW_hP'S'}$, where $\mathbf{W_h}$ is the variance-covariance matrix of the $h$-step ahead base forecast errors \cite{wickramasuriya2018optimal} . They showed that the $\mathbf{P}$ matrix that minimises the trace of $\mathbf{V}_h$ such that $\mathbf{SPS}=\mathbf{S}$ is given by $\mathbf{P}= ({\mathbf{S'} \mathbf{W}_h \mathbf{S}})^{-1} \mathbf{S'} {\mathbf{W}_h}^{-1}$. This will give the best (minimum variance) linear unbiased reconciled forecast, so called MinT approach. There are different methods to estimate $\mathbf{W}_h$. To do so, we use the shrinkage estimator and sample covariance as described in \cite{wickramasuriya2018optimal}. These models are shown with COM-SHR and COM-SAM, respectively.
	
	\section{Designing a dynamic hierarchical forecasting model} \label{dynamicproposalhier}
	
	Literature suggests designing hierarchies in a way that they are logical and have a high degree of homogeneity \cite{Fliedner1999,kahn1998revisiting}. In our investigated case, since retailers at the middle level have different sales patterns, some of the information may get lost if we aggregate them to find the total sales at the top-level \cite{Fliedner2001,gross1990disaggregation}. However, the bottom level series have a similar sales pattern to the middle-level series. This is because retailers at the middle level and DCs at the bottom level run promotion in the same period. Therefore, we use the MO approach to have a more homogeneous set of hierarchical time series. We forecast at the middle level (retailer's level). Then, we aggregate forecasts to the top level and disaggregate them to the bottom level. This also enables us to avoid information loss due to varying sales dynamic.
	
	The other important criterion that often has been missed in HF models is to use a dynamic model that evolves over time and considers time-series variations when aggregating or disaggregating series in the hierarchy. Compelling factors such as promotion can change the dynamic of demand and impact the efficacy of the HF models. Current static statistical models do not consider the sales variation caused by promotion over time. Therefore, we propose to use ML model for dynamic hierarchical forecasting (DHF).
	
	\subsection{Preprocessnig}
	Data preprocessing was required prior to modelling. We checked for missing values and possible outliers. We also checked for the seasonality of time series. Our investigated series do not have strong seasonality; however, they are highly impacted by promotion. We use the natural logarithm of sales to stabilise the variance. We also transferred the data using the min-max method for training the ANN model. 
	
	We train the models with actual time series values at the middle level and proportions of bottom level series. The input features include the sales time series of each parent node and the output is the proportions of corresponding children nodes. We train one model for each group of children-parent nodes. This ensures that the proportions of bottom level nodes sum up to one. Then, we forecast the parent node at the middle level and use the obtained forecasts as input values in ML models to predict the proportions of each child node over the forecasting horizon. We utilise three common ML models to dynamically disaggregate sales to the lower levels, whereby XGBoost, ANN, and SVR are used to forecast the proportions of the bottom level series from the middle level. We implemented the models in R programming language \cite{R}.   

	\subsection{XGBoost}
	
	XGboost is a decision tree based ensembling algorithm that uses a gradient boosting approach to generate unbiased and robust forecasts \cite{chen2016xgboost}. This algorithm uses a number of hyper-parameters. We tuned the parameters using a greedy search. We set the values of the learning rate, \textit{eta} on (0.01,0.05) interval by increment size of 0.01. The candidate values of column subsample and subsample size range from 0.3 to 1 by pace equal to 0.1.
	The values for the maximum number of boosting iterations rolls over the range of 100 to 500. The maximum depth of one tree is set between 2 and 10. We use a linear regression model as the objective function and choose the best results by minimising the root mean squared error. The optimal values of hyper-parameters vary for different investigated hierarchical time series. 
	
	\subsection{Artificial Neural Network} \label{annmodels}

	ANNs are powerful algorithms that are able to model any continuous, non-linear system, and can then make generalisations and predict the unseen values \cite{GuoqiangZhang1998}. There are many variations of ANN. We employed the commonly used `feed-forward error back-propagation' type of ANN with one hidden layer \cite{Carbonneau2008}. We used a greedy approach for optimising the hyper-parameters. We tested a sequence of different sizes by running the set size from 1 to 50 by increment size of 10. The learning rate decay was set on the sequence from 0.0 to 0.3 on a range of 0.1. We also tried both Sigmoid and Linear function and chose the linear function as the activation function. The model was fitted using the \textit{`neuralnet'} package in R \cite{gunther2010neuralnet}.

	\subsection{Support Vector Regression}\label{}
	SVR is a powerful supervised learning algorithm which is the most common application of support vector machines (SVM) \cite{Basak2007}. SVR finds a linear function in the space within a distance of $\epsilon$ from its predicted values. Any violation of this distance is penalised by a constant (C) \cite{Ali2009}. SVR employs a kernel function to transfer low dimensional data to high dimensional data.
	
	We applied greedy search to tune the hyper-parameters and choose the best combination of the kernel function, kernel coefficient $\gamma$, and cost function C. We tried different Kernel functions including poly, RBF, and linear. We ran $\gamma$ on a sequence of intervals of width 0.1 ranging from 0 to 1, and C on a sequence of intervals of width 1 ranging from 0 to 100 and fitted the best model using the `\textit{e1071}' package in R \cite{dimitriadou2006e1071}. 

	\section{Empirical results}\label{hierarchicalresults}
	
	This section summarises the empirical results driven with regards to the presented methods in Sections 4 and \ref{dynamicproposalhier}. We used ARIMAX with price as the explanatory variable to generate base forecasts at different levels. ARIMAX is an appropriate model to forecast since the demand time series are highly impacted by promotion. Forecasts are generated for one to eight steps ahead. We gathered data for 120 weeks of 61 hierarchical time series. We considered the first 112 weeks as the training-set and the last eights weeks as the test-set for validation purposes. We assess the forecasting accuracy using the sMAPE. The sMAPE is a scale-independent accuracy metric that can be used to compare the accuracy of models across different products and categories \cite{Hyndman2006}. 
	\begin{equation}
	\textit{sMAPE}= \frac{2}{n} \sum_{t=1}^{n} \frac{|Y_t -\hat{Y}_t|} {|Y_t + \hat{Y}_t|} \times 100\% , 
	\end{equation}
	where $n$ is the forecasting horizon, $Y_t$ is the actual value of the series, and $\hat{Y}_t$ is the generated forecast. We conducted ANOVA test to find out whether there is any significant differences between different models or not. We only report the results that shed light on our research methodology. Table \ref{hierarimaxmean} shows the results.

	\begin{table}[ht]
		\centering
		\caption{Forecasting accuracy of models: median sMAPE}
		\begin{tabular}{l|ccc|ccc|ccr}
			%		\toprule
			\hline
			\multicolumn{1}{l}{Horizon}&\multicolumn{3}{c}{One-step ahead}&\multicolumn{3}{c}{4-step ahead average}&\multicolumn{3}{c}{8-step ahead average}\\
			%		\midrule
			\hline
			\multicolumn{1}{l}{Level} & \multicolumn{1}{c}{Top}  &  \multicolumn{1}{c}{Middle} &  \multicolumn{1}{c}{Bottom} & \multicolumn{1}{c}{Top}  &  \multicolumn{1}{c}{Middle} &  \multicolumn{1}{c}{Bottom}  & \multicolumn{1}{c}{Top}  &  \multicolumn{1}{c}{Middle} &  \multicolumn{1}{c}{Bottom} \\
			\hline
			DHF-XGB & 18.87 & 25.65 & \textbf{25.54} & 16.99 & 26.60 & \textbf{26.26} & 18.64 & 28.96 & \textbf{29.02}  \\ 
			DHF-ANN & 18.87 & 25.65 & 27.07 & 16.99 & 26.60 & 29.00 & 18.64 & 28.82 & 31.78\\ 
			DHF-SVM & 18.87 & 25.65 & 25.69 & 16.99 & 26.60 & 26.78& 18.64 & 28.96 & 29.63 \\ 
			COM-SAM & 25.70 & 37.64 & 39.53& 23.95 & 43.03 & 45.65 & 23.85 & 42.90 & 46.29 \\ 
			COM-WLS & 17.25 & 25.17 & 26.59& 14.50 & 25.60 & 28.00 & 15.88 & 28.46 & 31.32\\ 
			COM-SHR & 17.58 & 25.02 & 26.80& {14.37} & \textbf{25.32} & 28.03 & 15.86 & \textbf{28.15} & 31.37\\ 
			TD & \textbf{16.81} & 29.47 & 31.68  & \textbf{14.36} & 28.86 & 31.53 & \textbf{15.80} & 31.51 & 34.47 \\ 
			TDFP & \textbf{16.81} & \textbf{24.44} & 27.07 & \textbf{14.36} & 26.08 & 28.78 &  \textbf{15.80} & 29.23 & 32.32\\ 
			CMO & 18.87 & 25.66 & 27.39& 16.99 & 26.60 & 29.08 & 18.64 & 28.95 & 32.09 \\ 		
			BU & 18.95 & 25.65 & 26.94& 16.29 & 25.53 & 28.05  & 18.55 & {28.17} & 31.36\\ 
			\hline
		\end{tabular}
		\label{hierarimaxmean}
	\end{table}

	The results indicate that the models' performances differ greatly across the hierarchy and depends on the forecasting horizon and the level of the hierarchy \cite{shlifer1979aggregation}.  In general, consistent with the literature we found that models accuracy is higher at the top level due to the aggregation \cite{rostami2015non}. Note that the ML models use the MO approach. Thus, they all generate the same forecasts at the middle and top levels. The conventional middle out (CMO) model in the Table \ref{hierarimaxmean} refers to the MO approach that uses the average of historical proportions for disaggregation.
	
	For the bottom level, the DHF-XGB model outperforms all other presented models including the BU model over the entire horizon. DHF-SVM model generates the second-best results. DHF-ANN depicts less accuracy than BU, COM-WLS, COM-SHR among statistical models; however, the difference is not significant. Other ANNs with a different architectures can be employed for further investigation. Nevertheless, this is a significant achievement for ML models that use the MO approach and outperform to BU that benefits the direct forecast at the bottom level. We found that there is a significant difference (at 5\% of significance level) between forecasting accuracy of ML models and state-of-art statistical models. Also, there is a significant difference among all utilised ML models at the bottom level. Our proposed model forecasts aggregated sales at the middle level that fits better to the structure of the investigated hierarchy and benefits from aggregation to improve the accuracy at the middle. Then, it uses ML models to dynamically disaggregate sales to the bottom level.

	For the middle level, while the TDFP generates the best results for the one-step-ahead forecast, the COM-SHR model generates more accurate forecasts for the four-step ahead and eight-step ahead horizons and is the second-best model in one-step ahead. The BU model generates a better forecast than the MO approach at the middle level, despite the fact that the MO directly forecasts the aggregated series at the middle level which has a similar pattern to the bottom level series. However, there is not a significant difference between forecasts of BU and MO approach at the middle level. 
	
	For the top level, the best forecasts across the entire horizon are generated with the TD and the TDFP models. The TD and TDFP models are the same at the top level as they both directly forecast the aggregated time series at the top level. Their good performance may attribute to the aggregation and the smaller variations of the total sales at the top-level \cite{rostami2015non}. The COM-SHR generates the second-best forecast at the top level which is not significantly different from TD and TDFP.
	
	Interestingly, the TD model slightly outperforms the BU model and MO models at the top level, even though the aggregated series at the middle level have different demand patterns and aggregation may cause information loss. That is, promotional information may get lost when aggregated to the top level. The obtained results show that TD models significantly outperform MO approaches despite different demand pattern of series at the middle level. This is in contrast with other results found in the literature \cite{kahn1998revisiting,rostami2015non}. This needs to be investigated further.

	\section{Conclusion}\label{hierarchicalconclusion}

	HF models can be used to improve the accuracy of time series at different levels. 
	Multiple time series that are nested can be modelled in a hierarchical fashion. The best HF model should be chosen based on demand behaviour, level of the forecasting, and the structure of the hierarchy. We considered a group of time series where the bottom level series are highly impacted by promotions and take different proportions of the sales at the middle level over time. We proposed to use ML models for dynamic proration. In this approach, we forecast sales at the middle level of the hierarchy and use ML models to forecast the proportions of the lower levels series from the middle level. We used XGBoost, ANN, and SVR as three common ML models in demand forecasting literature. 
	
	Our results show that the XGBoost outperforms other ML and statistical models at the bottom level and SVM generate the second-best results. Although we used XGBoost and SVM in a MO approach to indirectly forecast the bottom level and showed that it outperforms the BU model that directly forecasts at the bottom level. DHF-ANN also performed well and generated competitive forecasts to the other statistical models at the bottom level. This is a significant achievement that indicates the power of accurate disaggregation with ML models. 

	There is much debate in the literature about different HF models and their efficacy, and it is not trivial which one works better in different conditions \cite{syntetos2016supply}. One should choose the most appropriate model based on the desirable criteria and purpose of forecasting, i.e., the level of interest, the accuracy, and cost. This area of forecasting has received little attention from researchers in comparison to the other areas in the SC forecasting. For future research, we suggest to look at the correlation of demand series at different levels and embed the correlation in ML models. Our findings are based on data from the FMCG industry. It would be worthwhile to evaluate other data-sets from other industries. The other important problem is to consider the BU models with ML and investigate the promotion impact at different levels of the hierarchy to find the best HF model under the different promotional impact.
	
	\bibliographystyle{splncs04}
	\bibliography{library}
	
\end{document}